\documentclass[sigconf,screen=true,anonymous=false,bookmarks=false,nonacm]{acmart}
\usepackage{lipsum}
\usepackage{graphicx}
\usepackage{amsmath}
\usepackage{mathtools}
\usepackage{comment}
\usepackage[subrefformat=parens,labelformat=parens]{subfig}
\usepackage{bm}
\usepackage{multirow}
\usepackage{threeparttable,booktabs}
\usepackage{blkarray}
\usepackage{tikz}
\usetikzlibrary{positioning,calc,fit,decorations.pathmorphing,shapes.geometric,shapes.gates.logic.US,calc}
\usepackage{balance}
\usepackage{courier}                                       % courier font, used in \texttt
\usepackage{cleveref}                                      % smart citation
\Crefformat{figure}{Fig.~#2#1#3}                           % "Fig.", instead of "Figure"
\Crefname{subfigure}{Fig.}{Figs.}
\Crefname{figure}{Fig.}{Figs.}
\usepackage[mathcal]{eucal}
\usepackage[]{algpseudocode}                               % algorithm package
\algrenewcommand\textproc{\texttt}
\makeatletter\let\float@addtolists\relax\makeatother
\usepackage{algorithm}
\usepackage{filecontents}                                  % support to pgfplots
\usepackage{pgfplots}
\usepackage{pgfplotstable}
\pgfplotsset{compat=newest}
\usepackage[figuresright]{rotating}

\theoremstyle{plain}

\theoremstyle{definition}

\usepackage[skip=1pt]{caption}            % set space between figure and caption
\definecolor{CUHKorange}{RGB}{244,106,18} %F47012
\definecolor{CUHKblue}{RGB}{0,111,190}    %006FBE
\definecolor{CUHKgreen}{RGB}{0,127,128}   %007F80
\definecolor{CUHKred}{RGB}{228,46,36}     %E42E24
\definecolor{CUHKyellow}{RGB}{198,148,34} %C69422
\definecolor{CUHKdark}{RGB}{114,44,114}   %722C72
\definecolor{CUHKmiddle}{RGB}{144,44,144} %902C90

\usepackage{tcolorbox}
\tcbuselibrary{skins,breakable}
    {\endtcolorbox}

\usepackage{natbib}
\usepackage{graphicx}

\usepackage{enumitem}

\definecolor{bwwang}{rgb}{0,0,1}  
\definecolor{gbshen}{rgb}{0,1,0}
\definecolor{darkgreen}{rgb}{0,0.5,0}
\usepackage{adjustbox}
\usepackage{multirow}
\usepackage[english]{babel}
\title{LHNN: Lattice Hypergraph Neural Network for VLSI Congestion Prediction}

\author{Bowen Wang$^1$, Guibao Shen$^2$, Dong Li$^3$, Jianye Hao$^3$, Wulong Liu$^3$, Yu Huang$^4$, Hongzhong Wu$^4$, Yibo Lin$^5$, Guangyong Chen$^{2\star}$, Pheng Ann Heng$^1$}
\affiliation{
  \institution{$^1$Department of Computer Science and Engineering, The Chinese University of Hong Kong, $^2$Guangdong Provincial Key Laboratory of Computer Vision and Virtual Reality Technology, Shenzhen Institutes of Advanced Technology, \\$^3$Huawei Noah’s Ark Lab, $^4$Huawei Hisilicon, $^5$Department of Computer Science at Peking University}
  \country{}
  }
\email{bowenwang@link.cuhk.edu.hk}
\authornote{Corresponding author: Guangyong Chen} 

\begin{abstract}

Precise congestion prediction from a placement solution plays a crucial role in circuit placement. This work proposes the lattice hypergraph (LH-graph), a novel graph formulation for circuits, which preserves netlist data during the whole learning process, and enables the congestion information propagated geometrically and topologically. Base on the formulation, we further developed a heterogeneous graph neural network architecture LHNN,  jointing the routing demand regression to support the congestion spot classification. LHNN constantly achieves more than $35\%$ improvements compared with U-nets and Pix2Pix on the F1 score. We expect our work shall highlight essential procedures using machine learning for congestion prediction.
\end{abstract}

\begin{document}
\pagestyle{plain}

\maketitle

\section{Introduction}
\label{Introduction}

The global placement stage determines the fundamental physical layout of electronic cells on very-large-scale-integrated (VLSI) circuits, impacting the overall design performance while accounting for the major time consumption in the circuit design cycle. Given a placement solution, the global router in a routability-driven placement model is utilized to generate a congestion map, which is a binary mask identifying regions where routing demand exceeds routing capacity. The placer will accordingly optimize the correlated movable cell position to reduce both the wire length and the congestion, and repeat the above cycle iteratively until convergence to produce a placement solution. Employing a global router, e.g. NCTU-GR 2.0 \cite{liu2013nctu} can generate precise congestion maps. However, with the growth of circuit scale and complexity, time consumption tends to be unacceptable when utilizing a global router in the placement cycle to obtain the congestion map. In contrast, alternative methods for fast estimations of routing demand such as RUDY \cite{spindler2007RUDY} exist but are not reliable when identifying congestion areas.
 
 Recently, due to the need for the "shift-left" in circuit design, researchers begin to seek alternative solutions in machine learning \cite{tabrizi2018machine} \cite{chen2020pros} to achieve accurate and fast congestion map prediction. Current machine learning models commonly follow a two-phase workflow. First, based on domain knowledge, human experts generate various local features on the circuit using predefined functions on netlist. Then, based on the generated features, a specific model, e.g. convolution neural network (CNN) model is designed to predict either the routing demand map or the congestion map \cite{alawieh2020high-definition}. 
 
\begin{figure}[t!]
  \includegraphics[width=0.9\columnwidth]{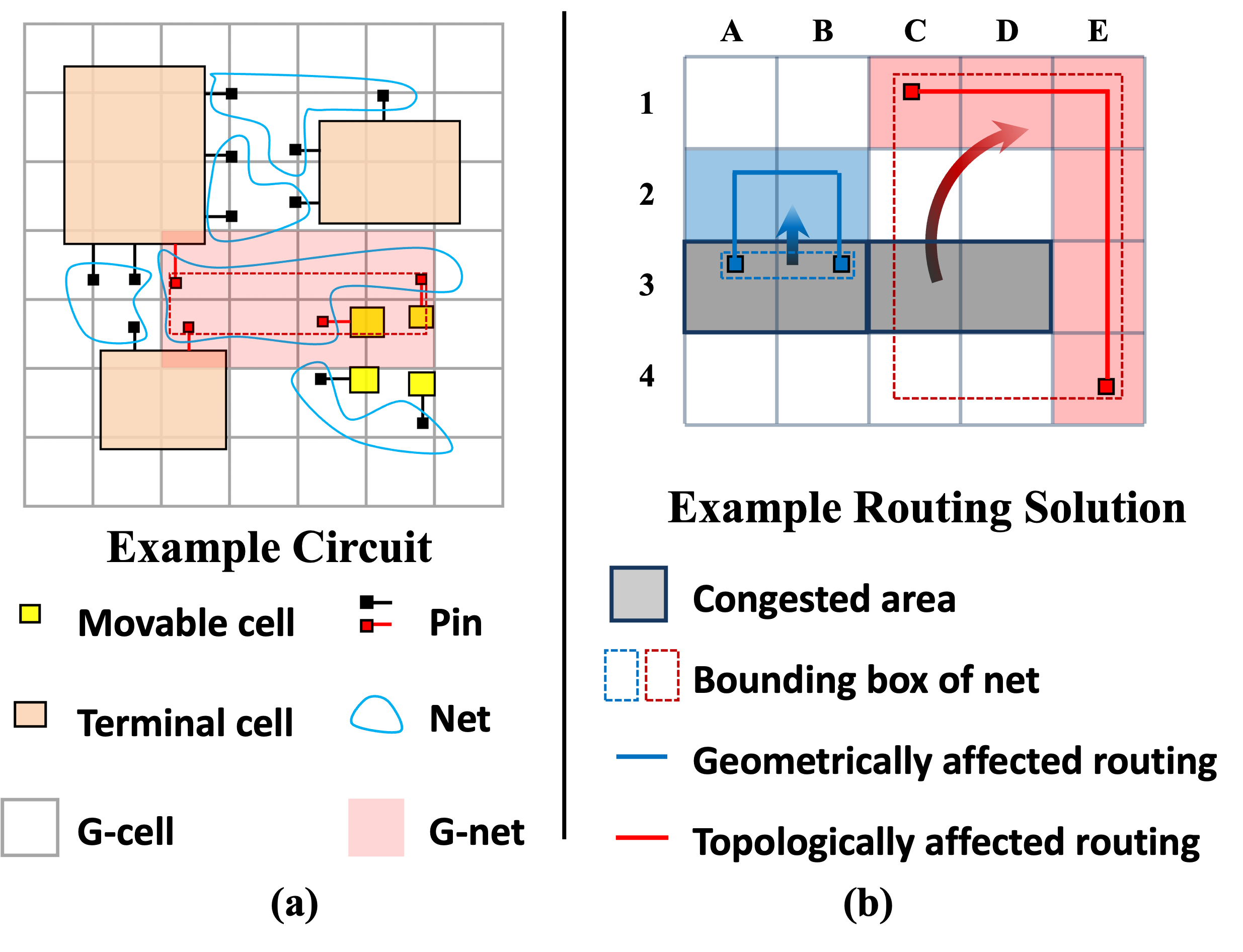}
  \caption{(a) Notations being used in this work. (b) Congestion condition of a G-cell influences others geometrically and topologically}

  \label{Notation}
\end{figure}

Several issues exist in current models. First, netlist on circuits is represented by converting net connections to crafted features on the local unit area. Critical netlist information can be overlooked and no longer available during this process. Moreover, routing demand and routing congestion are valuable complementary supervision signals, while current models only consider either of them as the learning target. Most importantly, the receptive field for CNN models is expanded purely on geometrical space, where a deep model is needed to capture the long-range connections. However, feature interactions between geometrically distant areas also commonly exist due to topological connections by nets. Figure \ref{Notation} (b) uses an example circuit with 2 nets to illustrate the two ways of feature interaction, supposing the 4 grid cells (G-cells) $\{3A,3B,3C,3D\}$ are congesting. For the blue net, knowing it is fully covered by congesting G-cells $\{3A,3B\}$, the best solution is to re-route its wire to geometrically nearest G-cells $\{2A,2B\}$ outside of the congested area. For the red net, observing that part of it is covered by congesting G-cells $\{3C,3D\}$, wires need to be routed to other positions according to the net coverage, namely detouring the congesting area while minimizing the total wire length. Therefore, wires are routed over $\{1C,1D,1E\}$ because these G-cells are topologically connected with the congested G-cells, even if geometrically distant. 

%  However, the features in the receptive field of conventional CNN-based models are either purely local or can only be propagated in geometric space, which hinders an accurate congestion prediction. 

\begin{figure*}[ht!]
  \includegraphics[width=\textwidth]{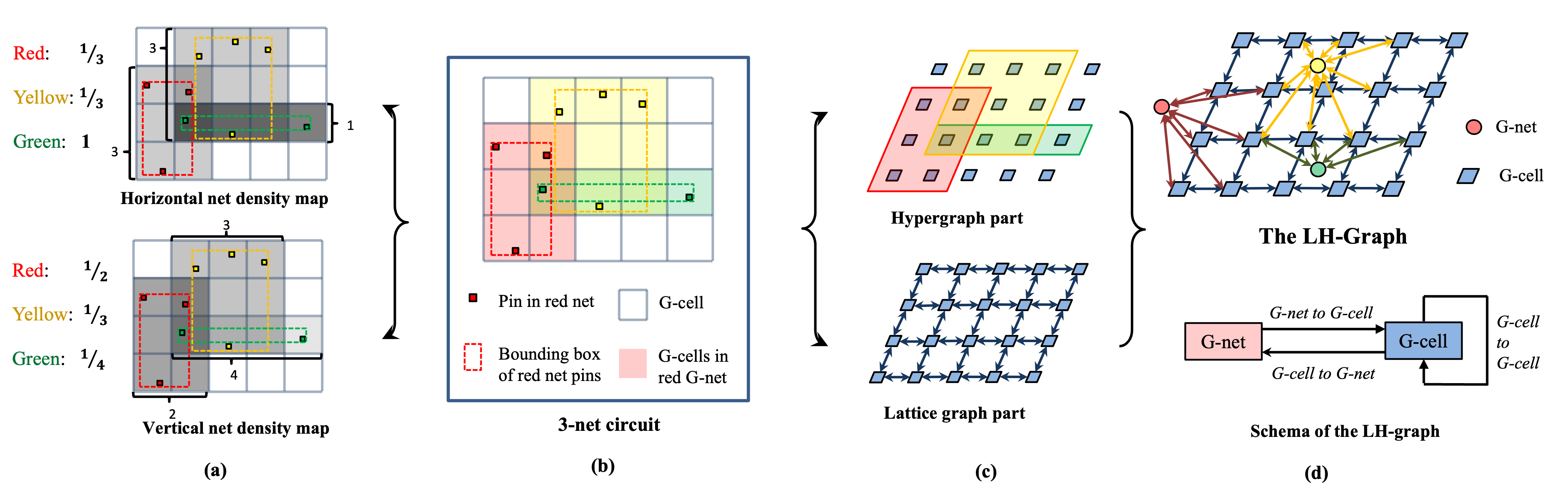}
    \setlength{\abovecaptionskip}{-10pt} 
  \caption{LH-Graph formulation and connection with crafted features. (a) Generation of net density maps. (b) A 3-net circuit example. (c) Hypergraph and lattice graph part of the circuit. (d) LH-graph and the correlated schema.}
  \setlength{\belowcaptionskip}{-4pt} 
  \label{Graph Formulation}
\end{figure*}

This work proposes to address the limitations induced by the current machine learning workflow. We target to reserve full netlist information during the whole learning process, fully utilize the supervision of demand map and congestion map, and facilitate more appropriate receptive field expansion. Our key contributions are summarized as follows: 

\begin{description}[style=multiline,labelsep=0cm,labelwidth=0.3cm,leftmargin=0.3cm,font=\normalfont]

\item[$\bullet$ ] We design the lattice hypergraph (LH-graph), a heterogeneous graph for VLSI circuits by combining hypergraph and lattice graph, which better preserves the netlist data while allowing the message to be propagated both in geometric space and in topological space. Most of the conventional features can be automatically generated during this process.

\item[$\bullet$ ] We propose Lattice Hypergraph Neural Network (LHNN), a heterogeneous graph neural network based on our graph formulation. By using multi-head supervision, two related tasks, i.e. routing demand regression and congestion classification, are jointly learned. 

\item[$\bullet$ ]We achieve more than $35\%$ improvements on F1 score over CNN models on the ISPD 2011 \& DAC 2012 challenge datasets. 

\end{description}

\section{Preliminaries and Related works}

\subsection{General Terms of VLSI Circuit}

We introduce some essential terms in VLSI using the example circuit shown in Figure \ref{Notation}(a). In this work, we generally distinguish movable cells with terminal cells, where the position of cells needs to be optimized or have been finalized during floor-planning, respectively. Pins are connectors on side of cells, and nets are a set of pins incident to a wire. Grid cells (G-cells) are rectangular units on circuits where cells are placed, usually represented as a pixel. 

We further introduce grid nets (G-nets), a set of G-cells that can fully contain the net coverage. Using the red net in Figure \ref{Notation} as an example, we first identify the bounding box of all pins, plotted as the dashed red rectangular. The respective G-net is the set of all 8 G-cells filled with red. Wires are most likely to be routed within the G-net to minimize the total wire length.

\subsection{Machine Learning Methods for Congestion Prediction}

Machine learning methods are drawing the attention of VLSI design automation researchers due to their ability to fast forecast possible congestion regions of a placement layout. After the early exploration of the machine learning methods \cite{tabrizi2018machine} \cite{hsu2014ntuplace4h}, due to the natural 2D planar structure of circuits, some CNN-based models \cite{xie2018routenet,chen2020pros,alawieh2020high-definition,al2021deep,J-Net} have been applied to the task. These models highly rely on hand-designed features based on domain knowledge, such as net density map, RUDY map, and pin density map \cite{alawieh2020high-definition}. Using crafted features could give a strong advantage for some models, while also putting a model at risk of losing important features from netlist data, which is no longer available in the latter learning process. Moreover, the receptive field of CNN models is expanded in geometrical space, where netlist-induced connections of different G-cells are completely missing.

More recently, the emergence of Graph Neural Network (GNN) triggered applications of undirected homogeneous graphs models on routing congestion prediction, since a VLSI circuit can be naturally represented by a graph. For example, CongestionNet \cite{kirby2019congestionnet} treats each cell as a node and net as an edge, then applies GAT to predict the routing demand on each cell based on netlist connection. \cite{chen2021detailed} treats each G-cell as a node, then applies GraphSAGE for the grid-like graph it formed for congestion prediction. These simple graph formulations cannot capture the geometrical and topological connections of circuits at the same time. Besides GNN architectures on simple graphs, various models operating on more advanced graph structures are recently under development. RGCN \cite{RGCN} operates on heterogeneous graphs which have different types of nodes and relations. HyperGCN \cite{yadati2018hypergcn} deals with hypergraphs, which can connect more than one node. The proposed model, LHNN is based on the RGCN framework and further developed by extending the concept of HyperGCN using the MPNN \cite{MPNN} pipeline.

\section{Graph Formulation and Analysis}
\label{Preliminaries}
\label{Formulation}
This section presents the formulation of the LH-graph, a heterogeneous graph for the VLSI circuit. We furthermore illustrate that commonly used features are automatically encoded in our graph formulation, and can be recovered simply by a one-step message passing through our graph structure. 

% \subsection{Motivation}
% \label{Motivation}

\subsection{Graph Formulation}

As illustrated in Figure \ref{Graph Formulation} (b) and (c), we propose to use a lattice graph to enable the message flow in geometric space. We regard each G-cell as a node and add an edge between two nodes if the respective two G-cells are adjacent. In the meantime, we use hypergraphs to facilitate the long-range information transfer in topological space. We regard each G-cell as a hypergraph node and treat each G-net as a hyperedge that connects all G-cells it contains. Finally, we use the heterogeneous graph to combine the hypergraph and the lattice graph, while preserving the representational space difference between the two types of graphs.

Formally, we formulate the circuit as LH-graph, a heterogeneous graph defined by $\mathcal{G}=(\mathcal{V}_c,\mathcal{V}_n,\mathcal{A},\mathcal{H})$ as Figure \ref{Graph Formulation}(d) shows. We define $\mathcal{V}_c \in \mathbb{R}^{\mathcal{N}_c \times d_c}$ and $\mathcal{V}_n \in \mathbb{R}^{\mathcal{N}_n \times d_n}$, where $\mathcal{N}_c$, $\mathcal{N}_n$ are the number of G-cells and G-nets respectively, with $d_c,d_n$ as the number of feature channels for G-cell and G-net. The $i$-th G-cell node feature is represented by $\mathcal{V}_{c[i,:]}$, i.e. the $i$-th row of $\mathcal{V}_c$. G-net node features are stored in the same way. Moreover, $\mathcal{A}\in\mathbb{R}^{\mathcal{N}_c \times \mathcal{N}_c}$ and $\mathcal{H}\in\mathbb{R}^{\mathcal{N}_c \times \mathcal{N}_n}$ are the adjacency matrix of lattice graph part and that of hypergraph part, respectively. Elements of the matrix $\mathcal{A}$ are defined as $\mathcal{A}_{ij}=1$ if two G-cells are adjacent to each other, and $\mathcal{A}_{ij}=0$ otherwise. Similarly, the elements of matrix $\mathcal{H}$ are defined as $\mathcal{H}_{ij}=1$ if G-cell $i$ is contained by G-net $j$, and $\mathcal{H}_{ij}=0$ otherwise. 

Based on the above definitions, we respectively denote the hypergraph part degree matrix for G-cell nodes and G-net nodes as diagonal matrices $\mathcal{D} \in \mathbb{R}^{\mathcal{N}_c \times \mathcal{N}_c}$ and $\mathcal{B}\in \mathbb{R}^{\mathcal{N}_n \times \mathcal{N}_n}$, where 
$\mathcal{D}_{ii} = \sum_{\epsilon=1}^{\mathcal{N}_n} \mathcal{H}_{i \epsilon}$ and 
$\mathcal{B}_{ii} = \sum_{\epsilon=1}^{\mathcal{N}_c} \mathcal{H}_{\epsilon i}$. Similarly, we define the lattice graph part degree matrix for G-cell nodes as $\mathcal{P}\in \mathbb{R}^{\mathcal{N}_c \times \mathcal{N}_c}$, where $\mathcal{P}_{ii} = \sum_{\epsilon=1}^{\mathcal{N}_c} \mathcal{A}_{i \epsilon}$.

The lower part of Figure \ref{Graph Formulation}(d) illustrates the schema of our constructed graph. To be specific, two types of vertices, \{ G-net, G-cell \}, are included in the heterogeneous graph. Three different types of relations are preserved among these nodes, which are \{ G-cell to G-net, G-net to G-cell, G-cell to G-cell \}. Thereby, nodes and edges of different types are allowed to be distinguishable in different representation spaces \cite{hu2020HGT}. 

In this work, we assign four channels of features to G-net nodes $\mathcal{V}_n$, which are $span_V$: the number of unit length the G-net cover vertically; $span_H$: the number of unit length the G-net cover horizontally; $npin$: the number of pins in the respective net; $Area$: the number of G-cells in G-net, equalling to $span_H \times span_V$. For G-cell features $\mathcal{V}_c$, following the definition in \cite{chen2020pros}, we include horizontal/vertical net density map, pin density map while adding the terminal cell mask: a binary mask on whether the G-cell is covered by a terminal cell. 

\subsection{Connection with Crafted Features}

With LH-graph, many effective features in CNN models can be recovered by a one-step message passing on G-net to G-cell relation. Take the net density map as an example, Figure \ref{Graph Formulation}(a) illustrates the generation process of this kind of feature. Net density map consists of two channels: horizontal/vertical net density, amount to the density of nets over each G-cell in horizontal/vertical direction. The feature is constructed based on the hypothesis that wires would be routed over each G-cell under an evenly distributed probability along the same direction. More specifically, the value of horizontal net density for each G-cell is generated by iterating each G-cell in each G-net and adding $\frac{1}{span_V}$ \cite{chen2020pros}. By assigning $\frac{1}{span_V}$ to each G-net node, the generation procedure of horizontal net density value on the G-cell node is equivalent to a one-step $sum$ aggregated message passing on G-net to G-cell relation.

Using similar principle, RUDY map can be approximated by adding $\frac{npin \times \left(span_H+span_V\right)}{Area}$ , and pin density map is in expectation equal to adding $\frac{npin}{Area}$. LH-graph can simply recover these two features by assigning $npin$, $span_V$, and $span_H$ to the G-net features, and use multilayer perceptron (MLP) to replace the aggregation function. Note that with multi-step message passing on LH-graph, more complex node embeddings that involve multi-step neighborhood features can be generated, attaining higher representational power. 

\section{The LHNN Architecture}
\label{LHNN}

\begin{figure*}[h!]
  \includegraphics[width=0.85\textwidth]{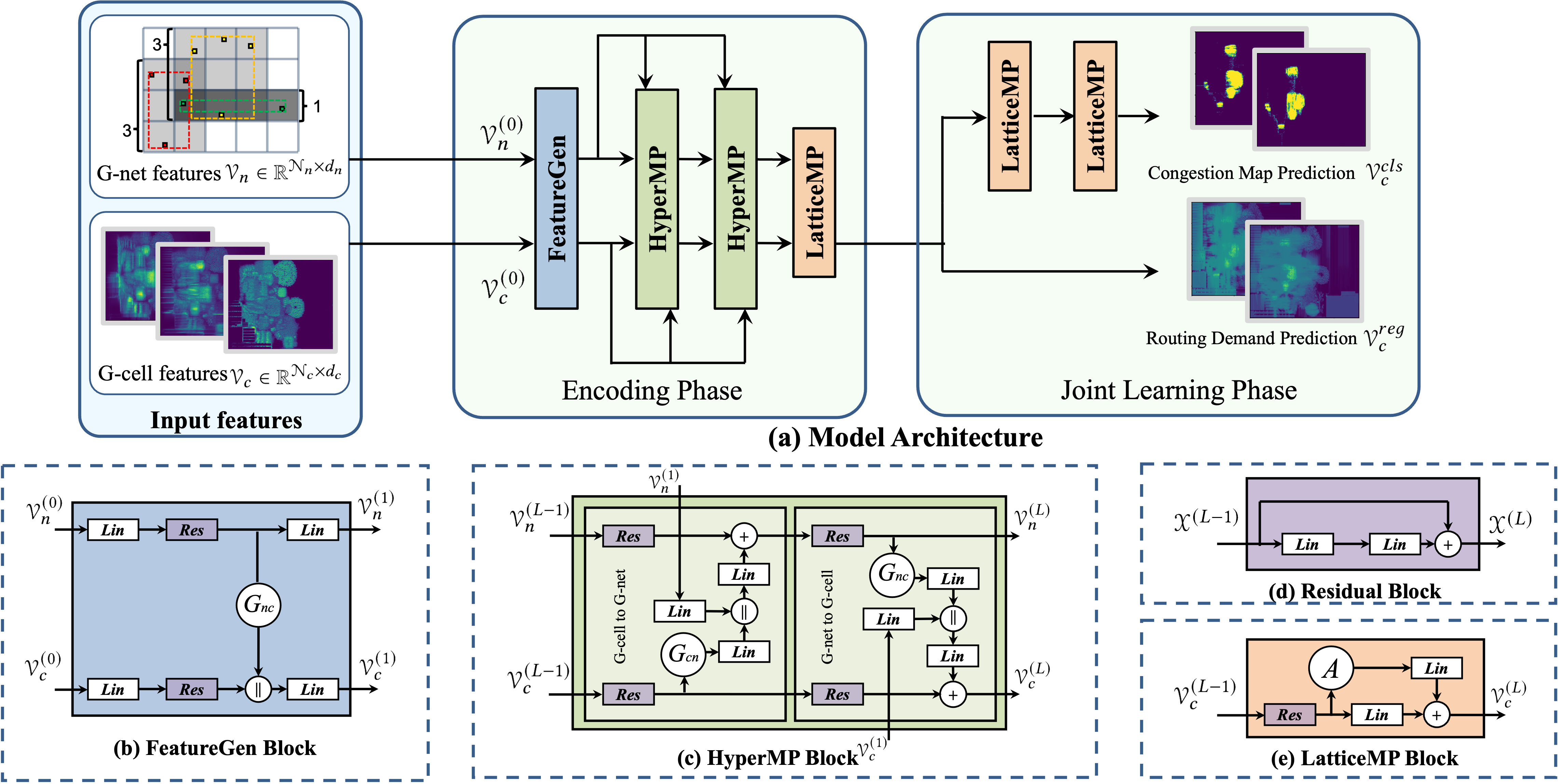}
  \caption{The LHNN architecture. $+$ denotes adding, $\|$ denotes concatenation, \textit{\textbf{Lin}} denotes linear layer $\mathbf{\sigma(W\mathcal{X}+b)}$ with learnable $\mathbf{w}$ and $\mathbf{b}$, \textit{\textbf{Res}} denotes residual block. The \textcolor{blue}{FeatureGen Block} takes $\mathcal{V}_c^{(0)}$ and $\mathcal{V}_n^{(0)}$ as input feature for G-cell and G-net, and generates the initial embedding vectors $\mathcal{V}_c^{(1)}$ and $\mathcal{V}_n^{(1)}$. The embedding vectors are further updated by multiple layers of \textcolor{darkgreen}{HyperMP Block}, which enables G-net and G-cell features to interact topologically. Finally, G-cell features will further proceed to stacks of \textcolor{orange}{LatticeMP Block}, where G-cell features are propagated geometrically. Routing demand signal is jointly supervising the model.}
  \label{Architecture}
\end{figure*}

% In this work, we propose HyperDGN, which stands for Hypergraph neural network combining directional message passing over lattice graphs. 
Based on the LH-graph, in this section, we present the LHNN architecture as shown in Figure \ref{Architecture}. The model is mainly composed of two phases: the encoding phase and the joint learning phase. Three main blocks: FeatureGen block, HyperMP block, and LatticeMP block are used to build the architecture. Messages will only be transferred on certain types of relations according to the block type. The encoding phase of LHNN will generate G-net and G-cell representations that preserve the underlying graph structure of circuits. The joint learning phase will joint the supervise signal from the routing demand map to support the congestion prediction.

\subsection{Feature Generation Block}

To ensure our model have the ability to replenish or transcend essential hand-crafted features, in Figure \ref{Architecture}(b), we deploy feature generation (FeatureGen) block as an analogy of the common feature generation process. We use MLPs with residual connections $\{f_{n}^{(1)}, f_{c}^{(1)}\}$ to transform input feature of G-cell and G-net to the embedding vectors of the hidden dimension. Next, we aggregate the G-net feature to its connected G-cells based on the G-net to G-cell relation $G_{nc}$, and fuse with the original transformed G-cell features with another linear layer. The process can be formally described as:

\begin{equation}
\mathcal{V}_c^{(1)} = \phi_{c}^{(1)} \left(f_{c}^{(1)}(\mathcal{V}_c^{(0)}) \| G_{nc}\left(f_{n}^{(1)}(\mathcal{V}_n^{(0)})\right) \right),
\end{equation}

\begin{equation}
\mathcal{V}_n^{(1)} = \phi_{n}^{(1)} \left(f_{n}^{(1)}(\mathcal{V}_n^{(0)})\right),
\end{equation}
where $G_{nc} = \mathcal{H}$, $\|$ denotes concatenation, $\{\phi_{n}^{(1)}, \phi_{c}^{(1)}\}$ represent the transformations after graph aggregation. 

\subsection{Hypergraph Message Passing Block}

 FeatureGen block is followed by stacks of hypergraph message passing (HyperMP) blocks and is carried out by operating among G-cell to G-net relation and G-net to G-cell relation alternately, shown in Figure \ref{Architecture}(c). The main purpose of this block is to expand the receptive field of each G-cell based on its netlist connection with others, preserve the connection during the feature encoding process. Intuitively, in each HyperMP block, the G-cell to G-net part helps each G-net to assess the routing demand condition in its covered area. The information will be processed by G-nets and distributed back to G-cells using the G-net to G-cell relation. Stacking layers of the block allows multi-step neighborhood feature fusion. Compared with CNN models, the HyperMP block enables topologically close G-cells to interact even if they are geometrically distant. 

More specifically, the block intakes four inputs: the inputs from the previous block $\mathcal{V}_c^{(L-1)}$, $\mathcal{V}_n^{(L-1)}$, and the feature generated by the FeatureGen block $\mathcal{V}_c^{(1)}$ and $\mathcal{V}_n^{(1)}$. In the G-cell to G-net part of the HyperMP block, we first use residual MLPs to transform the input features from the last layer. Then, using concatenation with a linear layer, we fuse the G-net feature generated by message aggregation on $G_{cn} = \mathcal{B}^{-1} \mathcal{H}^{\top}$ with G-net feature from FeatureGen block $\mathcal{V}_n^{(1)}$. Then, we add the result to the G-net feature transformed by the residual block from the previous layer. For the G-net to G-cell part of HyperMP block, we change $G_{cn}$ to $G_{nc}$ and conversely do the symmetric message propagation from G-net nodes to G-cell nodes.

\subsection{Lattice Graph Message Passing Block}

  We further added stacks of lattice graph message passing (LatticeMP) blocks, shown in Figure \ref{Architecture}(c). In HyperMP blocks, message passing is constrained by G-nets coverage, based on the assumption that wires will most likely be routed within the bounding box of nets. However, as illustrated in the blue net of Figure \ref{Notation}(b), information of routing demand can also be passed to geometric neighborhoods of the G-net when all its G-cells meet physical capacity. We therefore use the LatticeMP block to realize this message passing pattern. LatticeMP block operates solely on the G-cell to G-cell relation, where $A={\mathcal{P}}^{-1}\mathcal{A}$, along with a skip connection. 

\subsection{Jointing Supervision}

The necessity of placement optimization only exists in congested areas, rather than every G-cells with relatively high routing demand. However, simply using congestion maps as supervised labels can lead to sub-optimal performance. It is because that this procedure thresholds the real routing demand value on each G-cell into a binary value, leading to significant supervision information missing. In the meantime, simply thresholding predictions from the demand map regression model commonly result in an undesirable result, because the demand values between congestion areas and non-congestion areas are extremely close.

To improve the performance of congestion area prediction, we propose to preserve the supervision signal from routing demand map via the joint learning phase as Figure \ref{Architecture} shows. Let $\mathcal{V}_c^{cls}=\left\{\mathbf{c}^{cls}_{i}\right\}_{i=1}^{\mathcal{N}_{c}}$ and  $\mathcal{V}_c^{reg}=\left\{\mathbf{c}^{reg}_{i}\right\}_{i=1}^{\mathcal{N}_{c}}$ be the sets of predictions of congestion condition and demand value for each G-cell respectively, with subscript $n$ representing the $n$-th G-cell in training set. Let $Y_{cls}=\left\{\mathbf{y}^{cls}_{i}\right\}_{i=1}^{\mathcal{N}_{c}}$ and $Y_{reg}=\left\{\mathbf{y}^{reg}_{i}\right\}_{i=1}^{\mathcal{N}_{c}}$ be the ground truth of congestion map and routing demand map. We equally treat the two tasks by setting the objective loss function $\mathcal{L}$ as:

\begin{equation}
\begin{gathered}
\mathcal{L}=\mathcal{L}_{reg}+\mathcal{L}_{cls},
\end{gathered}
\end{equation}

\begin{equation}
\label{reg}
\begin{gathered}
\mathcal{L}_{reg}=-\frac{1}{\mathcal{N}_{c}}\sum_{i=1}^{\mathcal{N}_{c}}\left(\mathbf{c}^{reg}_{i}-\mathbf{y}^{reg}_{i}\right)^2,
\end{gathered}
\end{equation}

\begin{equation}
\label{cls}
\begin{gathered}
% \allowdisplaybreaks
\mathcal{L}_{cls}=
-\frac{1}{\mathcal{N}_{c}} \sum_{i=1}^{\mathcal{N}_{c}} 
\Bigg\{\left[ \left(1-\mathbf{y}^{cls}_{i}\right) \gamma + \mathbf{y}^{cls}_{i} \right] \times \\
\left[\mathbf{y}^{cls}_{i}  \log \mathbf{c}^{cls}_{i} + 
\left(1-\mathbf{y}^{cls}_{i}\right) \log \left(1-\mathbf{c}^{cls}_{i}\right)\right]\Bigg\},
\end{gathered}
\end{equation}
where $\mathcal{L}_{reg}$ and $\mathcal{L}_{cls}$ denotes regression and classification loss, using Mean Square Error (MSE) loss and Binary Cross Entropy (BCE) loss, respectively. Empirically, the congestion classification task suffers a lot from label imbalance. Therefore, we add a hyperparameter $\gamma \in (0,1]$ to reduce the loss of each non-congesting G-cells, relieving the tendency of predicting each G-cells as non-congested. We thereby achieve better performance than solely relying on the congestion map by preserving both supervision signals.

\section{Experimental Results}
\label{Experiments}
% In this section, we used real-world datasets to evaluate the effectiveness of our LH-graph formulation and LHNN model.

\subsection{Experimental Settings}

\textbf{Dataset.}
We conduct experiments on ISPD 2011 \cite{ispd2011dataset} and DAC 2012 \cite{dac2012dataset} contest benchmarks, and have 15 different VLSI designs in total. We use 10 for training and 5 for testing, and run DREAMPlace \cite{lin2020dreamplace} on each of the designs to generate placement solutions. Then, we further apply NCTU-GR 2.0 \cite{liu2013nctu} to attain horizontal/vertical routing demand maps, and set the congestion maps as a binary indicator according to whether the horizontal/vertical routing demand of the G-cell exceeds the circuit's capacity. These two kinds of maps are regarded as labels in Eq.\ref{reg} and \ref{cls}, respectively.

We further notice that experiments on the random splitting of the dataset lead to extremely large performance variation, because the average congestion condition of the training set may largely differ from the testing set.
% Besides, we notice that random split settings lead to extremely ambiguous results. This is caused by the large variation of the congestion rate of different circuits. When the mean congestion rate of the training set and testing set largely differs, the domain transfer effect is dominating the overall performance variation. 
To avoid the ambiguity caused by the domain transfer effect, we iterate all 10:5 splits and fix our split that minimizes the average congestion rate difference between the training set and testing set. Finally, the average ratio of congested G-cell is $17.38\%$ in both sets. We use the same split for all experiments, and summarize the dataset statistics in Table \ref{Dataset information}. 
% Note that the design cases for testing are completely unseen during the training phase.
%now 17.38

\begin{table}[]
  \caption{Dataset Information}
  \label{Dataset information}
  \centering
\begin{adjustbox}{width=\columnwidth,center}
\begin{tabular}{|c|c|ccc|c|}
\hline
Split                     & Superblue            & \multicolumn{3}{c|}{Data Info Average}                                                                          & Congestion             \\ \cline{3-6} 
                          & Circuit ID           & \multicolumn{1}{c|}{\#cells}               & \multicolumn{1}{c|}{\#nets}                & \#G-cells             & rate (\%)              \\ \hline
\multirow{2}{*}{Training} & 2,3,4,7,10           & \multicolumn{1}{c|}{\multirow{2}{*}{866K}} & \multicolumn{1}{c|}{\multirow{2}{*}{847K}} & \multirow{2}{*}{311K} & \multirow{2}{*}{17.38} \\
                          & 12,14,16,18,19       & \multicolumn{1}{c|}{}                      & \multicolumn{1}{c|}{}                      &                       &                        \\ \hline
Testing                   & 1,5,6,9,11           & \multicolumn{1}{c|}{887K}                  & \multicolumn{1}{c|}{877K}                  & 406K                 & 17.38                  \\ \hline
\textbf{Total}            & \textbf{All designs} & \multicolumn{1}{c|}{\textbf{873K}}         & \multicolumn{1}{c|}{\textbf{857K}}         & \textbf{343K}         & \textbf{17.38}         \\ \hline
\end{tabular}
\end{adjustbox}
\end{table}

\textbf{Evaluations metrics.}
 We use two primary metrics F1 score (F1) and accuracy (ACC) to evaluate model performances. For all models, we fix the number of training epochs and repeat each experiment under different random seeds for $5$ times. We report the average and standard deviation of two metrics on the final epoch over the testing set. Note that some circuits can have zero congestion rate, which F1 score will always turn out to be zero, holding back the average F1 score results.

\textbf{Models Configurations.}
We implement our model using the Deep Graph Library(DGL) \cite{wang2019dgl} in PyTorch with an NVIDIA Tesla T4 GPU. Our model applies 2 layers of stacking HyperMP block and 1 layer of LatticeMP block for the encoding phase, and another 2 LatticeMP blocks in the joint learning phase.
% For the encoding phase, our model applies 2 layers of stacking HyperMP block and one layer of LatticeMP block. For the classification task in the jointing Phase, our model further deployed 2 LatticeMP blocks before congestion prediction.
We use 32 as the hidden dimension throughout the whole model. To save computational cost, we applied mini-batch training following the instructions of DGL, and use a random sampler in each aggregation. The number for neighbor sampling are \{6,3,2\} for FeatureGen, HyperMP, LatticeMP block respectively. To avoid the neighborhood sampling process being dominated by the large degree G-nets, for each circuit, we further removed all large G-nets that contain more than $0.25\%$ of the total G-cell number of the circuit. We used the Adam optimizer with a learning rate of $2e^{-3}$ and $5e^{-4}$ to train LHNN. For all experiments, we set $\gamma$ to be 0.7 to reduce the label imbalance effect.

\subsection{Model Comparison and Visualization}

\begin{table}[t!]
  \caption{Model comparison results on ISPD 2011 \& DAC 2012}
  \label{Comparison}
  \centering
\begin{adjustbox}{width=\columnwidth,center}
\begin{tabular}{|c|cc|cc|}
\hline
\multirow{2}{*}{Model} & \multicolumn{2}{c|}{Uni-channel}                               & \multicolumn{2}{c|}{Duo-channel}                                  \\ \cline{2-5} 
                       & \multicolumn{1}{c|}{F1 score}              & ACC                   & \multicolumn{1}{c|}{F1 score}              & ACC                   \\ \hline
4-layer MLP            & \multicolumn{1}{c|}{32.58$\pm$0.37}          & 94.29$\pm$0.13          & \multicolumn{1}{c|}{28.95$\pm$0.64}          & 94.27$\pm$0.10          \\ \hline
Pix2Pix                    & \multicolumn{1}{c|}{30.20$\pm$0.53}                    & 93.82$\pm$0.26                    & \multicolumn{1}{c|}{28.31$\pm$0.54}                    & 93.13$\pm$0.13                    \\ \hline
U-net                  & \multicolumn{1}{c|}{29.75$\pm$3.03}          & 94.45$\pm$0.19          & \multicolumn{1}{c|}{29.52$\pm$3.27}          & 92.28$\pm$0.45          \\ \hline
\textbf{LHNN(Ours)}    & \multicolumn{1}{c|}{\textbf{40.89$\pm$1.82}} & \textbf{95.46$\pm$0.11} & \multicolumn{1}{c|}{\textbf{37.48$\pm$2.34}} & \textbf{94.77$\pm$0.13} \\ \hline
\end{tabular}
\end{adjustbox}
\end{table}

\begin{figure}[t!]
  \includegraphics[width=\columnwidth]{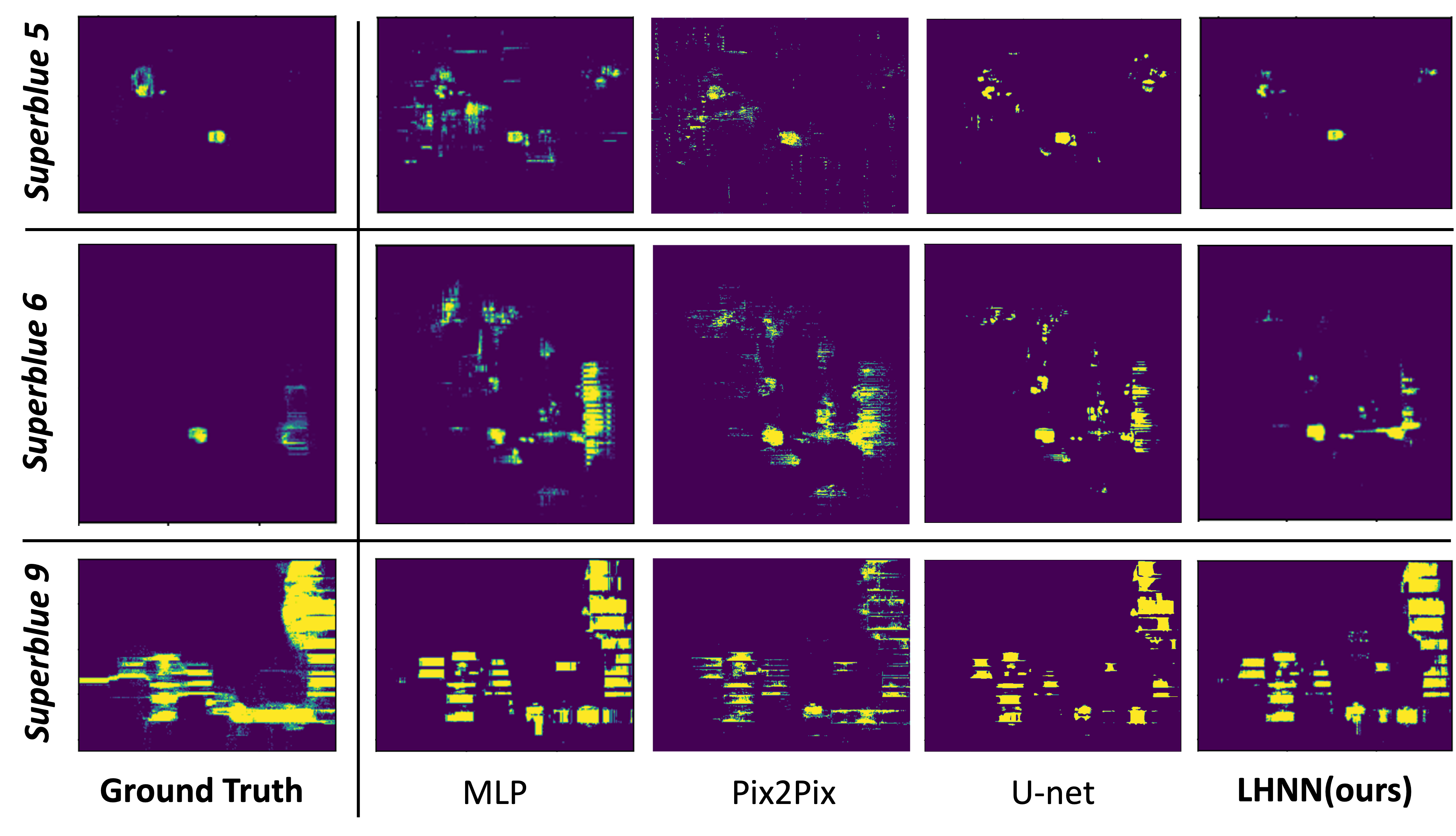}
  \caption{Visualization of model performance comparison.}
  \label{Visualization}
\end{figure}

 To evaluate our model performance while discover the true factor that affects the congestion prediction performance, we implement 3 conventional models for comparison. To assess the effectiveness of local crafted features, we formulate a conventional 4-layer MLP with residual connection for each G-cell as the vanilla baseline, using common hyper-parameters as LHNN. To validate the importance of receptive field expansion in topological connection, we further compare our model with U-net\cite{ronneberger2015U-net} and Pix2Pix\cite{isola2017pix2pix} using the top PyTorch implementations in Github \footnote{https://github.com/milesial/Pytorch-UNet   and https://github.com/junyanz/pytorch-CycleGAN-and-pix2pix}, which are widely acknowledged effective models in pixel-wise classification tasks. For U-Net and Pix2Pix, $256\times256$ crop size is adopted during training and testing. We treat G-cell features as the four-channel input for each model, and use congestion mask as the prediction label, applying the label balance factor $\gamma$ the same as LHNN.

The experimental results are listed in Table \ref{Comparison}. Uni-channel indicates that we are solely predicting horizontal congestion map, and duo-channel refers to predicting horizontal/vertical congestion maps at the same time. We first notice that our LHNN significantly outperforms all baseline models in all metrics, for both uni-channel and duo-channel tasks. The F1-score of LHNN achieves astonishing 37.44\% improvements over U-net, and 35.39\% over Pix2Pix for the uni-channel experiment, indicating the effectiveness of our graph formulation and architecture. Furthermore, we notice that hand-crafted features are sufficiently informative, where simple MLP that operates on completely local features can already achieve comparable performance with U-net. The result indicates that solely relying on geometrical message passing while abandoning the topological connections induced by netlist is not sufficient for the congestion prediction task. We visualize the uni-channel prediction results for Superblue 5, 6, and 9 in Figure \ref{Visualization}, whose congestion rates are 1.43\%, 1.13\%, and 47.70\% respectively. We notice that LHNN manifest to distinguish circuits of different congestion level, while conventional models tend to generate a more averaged congestion condition among different circuits, increasing false positives in low congestion circuits and false negatives in high congestion circuits.

\begin{table}[]
  \caption{Ablation study on uni-channel experiments}
  \label{Ablation}
  \centering
  \begin{adjustbox}{width=\columnwidth,center}
\begin{tabular}{|c|cccccc|}
\hline
Module Name                        & \multicolumn{6}{c|}{Model}                                                                                                                                                           \\ \hline \hline
FeatureGen                    & \multicolumn{1}{c|}{\checkmark} & \multicolumn{1}{c|}{}           & \multicolumn{1}{c|}{\checkmark} & \multicolumn{1}{c|}{\checkmark} & \multicolumn{1}{c|}{\checkmark} & \checkmark \\ \hline
HyperMP                          & \multicolumn{1}{c|}{\checkmark} & \multicolumn{1}{c|}{\checkmark} & \multicolumn{1}{c|}{}           & \multicolumn{1}{c|}{\checkmark} & \multicolumn{1}{c|}{\checkmark} & \checkmark \\ \hline
LatticeMP                          & \multicolumn{1}{c|}{\checkmark} & \multicolumn{1}{c|}{\checkmark} & \multicolumn{1}{c|}{\checkmark} & \multicolumn{1}{c|}{}           & \multicolumn{1}{c|}{\checkmark} & \checkmark \\ \hline
Jointing                          & \multicolumn{1}{c|}{\checkmark} & \multicolumn{1}{c|}{\checkmark} & \multicolumn{1}{c|}{\checkmark} & \multicolumn{1}{c|}{\checkmark} & \multicolumn{1}{c|}{}           & \checkmark \\ \hline
G-cell Feature                     & \multicolumn{1}{c|}{\checkmark} & \multicolumn{1}{c|}{\checkmark} & \multicolumn{1}{c|}{\checkmark} & \multicolumn{1}{c|}{\checkmark} & \multicolumn{1}{c|}{\checkmark} &            \\ \hline \hline
F1                                 & \multicolumn{1}{c|}{\textbf{40.89}}      & \multicolumn{1}{c|}{38.99}      & \multicolumn{1}{c|}{32.53}      & \multicolumn{1}{c|}{36.52}      & \multicolumn{1}{c|}{35.72}      & 38.02      \\ \hline
$\frac{\Delta F1}{F1_{full}}$ (\%) & \multicolumn{1}{c|}{\textbf{0}}          & \multicolumn{1}{c|}{-4.65}      & \multicolumn{1}{c|}{-20.45}     & \multicolumn{1}{c|}{-10.69}     & \multicolumn{1}{c|}{-12.64}     & -7.02      \\ \hline
\end{tabular}
\end{adjustbox}
\end{table}

\subsection{Ablation Studies}

% \usepackage{lscape}
% \begin{landscape}

To further confirm the effectiveness of all our components, we perform ablation studies based on the uni-channel task. The top part indicates the existence of our key components. If no tick in FeatureGen, HyperMP, LatticeMP module, we remove every edge of correlated relation types in the respective block, verifying the significance of topological and geometrical message passing. The linear and residual layer in the block is preserved to keep the depth and parameter number of the model approximately the same. No ticking in the jointing module indicates the removal of the entire regression branch, validating the effectivity of routing demand map supervision. No ticking in the G-cell feature module indicates setting the pin density, net density channels in G-cell to be zero, challenging our argument that some G-cell features can be recovered or enhanced.

From the results, we notice that removing HyperMP block edges leads to the most drastic downfall of model performance, where most topological connection within G-cells exists within this block. Removing LatticeMP block edges also leads to a significant F1 score reduction. These results prove our argument that the receptive field needs to be expanded via topological space and horizontal space. Secondly, removing the regression branch can also largely drop the model performance, providing evidence that the demand map contains important supervise signals to our model that cannot be ignored. Finally, we notice that even when we only keep the terminal mask feature in the G-cell, our model can still achieve $38.02\pm4.57$ in F1 score, while MLP, U-net, and Pix2Pix can completely fail in this setting. This affirms our statement that the removed G-cell features can be automatically recovered by our graph formulation.

\section{Conclusion}
\label{Conclusion}
In this work, we formulate a novel heterogeneous graph structure LH-graph for VLSI routing congestion prediction during the placement stage. Based on the formulated graph, we develop LHNN, a model that can perform message passing between G-cells both in geometrical space and in topological space, and jointly learn the routing demand regression and congestion classification. We validate the effectiveness of the proposed key components by comparing experiments and ablation studies on ISPD 2011 and DAC 2012 datasets. LHNN significantly outperforms conventional methods that solely rely on local or geometrical space message passing, and is much benefited from jointly learning two labels. We believe our work reveals crucial behaviors to be noticed in the development of future machine learning models for routability-driven placement.

\section*{Acknowlegment}
\label{Acknowledgement}
The work is supported by the National Key Research and Development Program of China(2020YFB1313900), Guangdong Provincial Basic and Applied Basic Research Fund - Regional Joint Fund (2020B1515130004), the National Natural Science Foundation of China (Project No. 62006219) and the Key Fundamental Research Program of Shenzhen (JCYJ20210324115601004).

% \printbibliography
\bibliography{DACRef}
% \bibitem{b1} G. Eason, B. Noble, and I. N. Sneddon, ``On certain integrals of Lipschitz-Hankel type involving products of Bessel functions,'' Phil. Trans. Roy. Soc. London, vol. A247, pp. 529--551, April 1955.

% \end{thebibliography}
\vspace{12pt}

\end{document}